\documentclass[10pt,twocolumn,letterpaper]{article}

\usepackage{cvpr}
\usepackage{times}
\usepackage{epsfig}
\usepackage{graphicx}
\usepackage{amsmath}
\usepackage{amssymb}
\usepackage{amsthm} 
\usepackage{algorithm}
\usepackage{algorithmic}
\usepackage{booktabs}
\usepackage{longtable}
\usepackage{multirow}
\usepackage{subfigure}
\usepackage{caption}

\usepackage[breaklinks=true,bookmarks=false]{hyperref}

\cvprfinalcopy 

\def\cvprPaperID{xxxx} 

\def\beq{\begin{equation}}
\def\eeq{\end{equation}}

\newcommand{\bbb}[1]{\boldsymbol{\mathbf{#1}}}

\begin{document}

\title{Mutually-aware Sub-Graphs Differentiable Architecture Search}

\author{{Haoxian Tan$^{1}$} $\,\,\,\,\,\,\,$ Sheng Guo$^{2}$ $\,\,\,\,\,\,\,$ Yujie Zhong$^{1}$ $\,\,\,\,\,\,\,$ Matthew R.Scott $^{1}$ $\,\,\,\,\,\,\,$ Weilin Huang$^{1}$ \\

$^{1}$Malong LLC. $\,\,\,\,\,\,$  $^{2}$MYBank, Ant Group\\

{\tt\small tanhx3@mail3.sysu.edu.cn} $\,\,\,$ {\tt\small guosheng.guosheng@alibaba-inc.com} \\ {\tt\small jasper901013@hotmail.com} $\,\,\,$
{\tt\small mscott@malongtech.com} $\,\,\,$ {\tt\small weilin\_h@hotmail.com}

}

\maketitle
\begin{abstract}
Differentiable architecture search is prevalent in the field of NAS because of its simplicity and efficiency, where two paradigms, multi-path algorithms and single-path methods, are dominated.
Multi-path framework (e.g.$~$DARTS) is intuitive but suffers from memory usage and training collapse. 
Single-path methods (e.g.$~$GDAS and ProxylessNAS) mitigate the memory issue and shrink the gap between searching and evaluation but sacrifice the performance. 
In this paper, we propose a conceptually simple yet efficient method to bridge these two paradigms, referred as Mutually-aware Sub-Graphs Differentiable Architecture Search (MSG-DAS).
The core of our framework is a differentiable Gumbel-TopK sampler that produces multiple mutually exclusive single-path sub-graphs. 
To alleviate the severer \textit{skip-connect} issue brought by multiple sub-graphs setting, we propose a Dropblock-Identity module to stabilize the optimization.
To make best use of the available models (super-net and sub-graphs), we introduce a memory-efficient super-net guidance distillation to improve training.
The proposed framework strikes a balance between flexible memory usage and searching quality. 
We demonstrate the effectiveness of our methods on ImageNet and CIFAR10, where the searched models show a comparable performance as the most recent approaches.
\end{abstract}

\section{Introduction}
Deep neural networks have achieved great success in computer vision in the recent years, and great human effort has been continuously devoted to design a deeper, stronger, and more powerful neural architecture, starting from AlexNet~\cite{krizhevsky2012imagenet}, VGGNet~\cite{simonyan2014very}, GoogleNet~\cite{szegedy2015going}, to recent ResNet~\cite{he2016deep} and SENet~\cite{hu2018squeeze}.
However, the unlimited choices of network architectures make manual search infeasible and impractical. In the past two years, Neural Architecture Search (NAS) has gained increasing attention. It explores a large search space of neural architectures and is able to automatically discover a high-performance architecture by using various searching algorithms, such as evolution method~\cite{real2019regularized} or reinforcement learning~\cite{zoph2018learning}. Recently, a number of works followed this line of research, such as \cite{zoph2016neural,liu2018darts,Nekrasov2018FastNA,Ghiasi2019NASFPNLS,Wang2019NASFCOSFN}, which have proved that NAS-based approaches can reduce the labor of human experts significantly on the design of neural networks, and are able to find better neural architectures for various computer vision tasks.

However, early architecture search algorithms are computationally expensive despite their promising performance. They often take many thousands GPU days to search a state-of-the-art architecture, by using reinforcement learning (RL)~\cite{zoph2016neural,zoph2018learning,pham2018efficient} or evolution algorithm (EA)~\cite{real2017large,elsken2018efficient,real2019regularized,liu2017hierarchical}. Recent works attempted to speed up the searching process by using weight sharing technique~\cite{pham2018efficient,bender2018understanding,guo2019single,liu2018darts, Zhong2020RepresentationSF, Liu2021UnchainTS}.

\begin{figure}[tb]
\centering
\includegraphics[trim=160 200 380 220, clip,height=4cm]{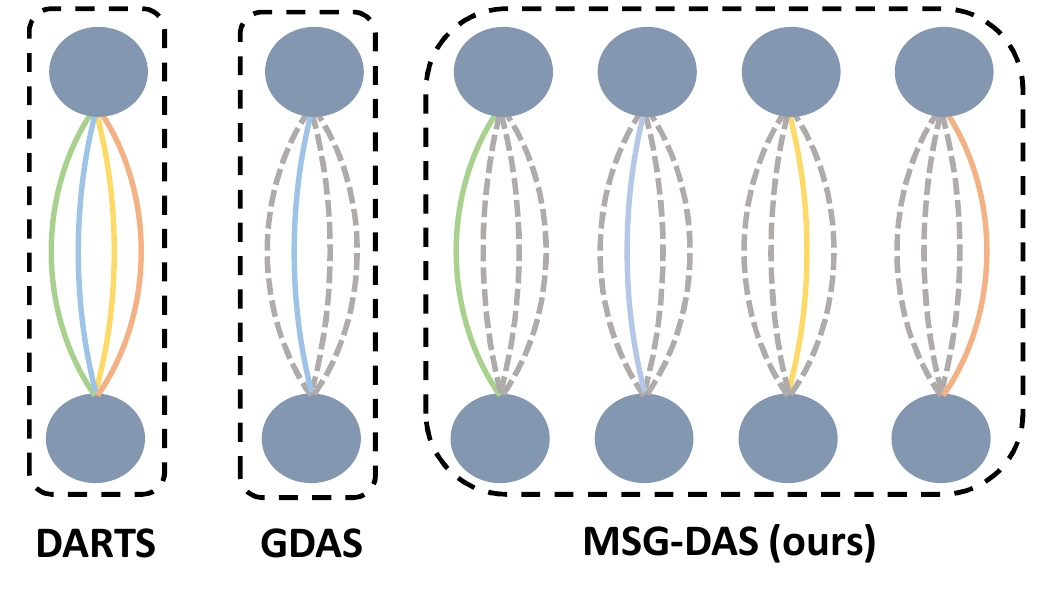}
\captionsetup{font=small}
\caption{\textbf{Visual comparison between DARTS, GDAS and our MSG-DAS.} All candidate operations are optimized simultaneously in DARTS while only one operation is considered in GDAS. MSG-DAS optimizes $K$ single-path 
sub-graphs whose edges sample different operations.}
\label{fig:compare}
\vspace{-0.1cm}
\end{figure}

The weights of all candidate operations in DARTS~\cite{liu2018darts} are learned simultaneously during training process, but only one optimal operation is selected for the derived network architecture. 
This comes up with a few limitations on DARTS. 
First, all weights in the super-net are learned simultaneously at each training iteration, which explore a large amount of memory cost.

Second, the training of whole super-net easily leads to Matthew effect~\cite{merton1968matthew}. Some routes can be strengthened in the early stage of the training, which may compress the optimization space of the other routes. Thus the search may fall into a local optima in the early stage.
Third, as indicated in \cite{liang2019darts+,zela2019understanding}, the architecture searched by DARTS often results in proliferation of skip-connect operation, leading to a significant reduction of model parameters. This inevitably limits the representative power of the derived networks, thus producing unsatisfied results. 

Recently, ProxylessNAS~\cite{cai2018proxylessnas} and GDAS~\cite{dong2019searching} were proposed by designing a single-path sampling method at each iteration to narrow the gap between searching and retraining. 
 
The two methods not only reduce memory usage, but also provide a more accurate approximation of the derived model. However, they can be viewed as a random block coordinate descent algorithm~\cite{nesterov2012efficiency}, which may magnify the local-optima issue for a non-convex problem due to unstable sampling.

In this work, we study both multi-path (e.g., DARTS) and singe-path approaches (e.g., GDAS), and propose Mutually-aware Sub-Graphs Differentiable Architecture Search (MSG-DAS), which is able to train multiple exclusive single-path sub-graphs by exploring the joint merits of both singe-path and multi-path methods.
Our MSG-DAS updates the super-net at the path level like other single-path methods which guarantee the consistency between the searching model and derived model (Figure \ref{fig:compare}). However, our method can alter the number of single-path sub-graph according to the given resource, and more weights can be updated at each iteration like multiple-path methods. To cover the super-net to a larger extent, and at the same time, to avoid updating the same weights multiple times in a single iteration, these sub-graphs do not share any operation. In other words, they are mutually exclusive. The sampling of mutually exclusive sub-graphs can be implemented in a differentiable manner by our Gumbel TopK sampler.
 
As mentioned in \cite{liang2019darts+, chen2019progressive}, a key limitation of DARTS is the spread of skip-connect operation, which directly leads to a significant shrinkage of the searched cell. This naturally limits the capacity of the derived networks, thus producing frustrated results. In this work, we propose to replace the regular Dropout  with Dropblock~\cite{ghiasi2018dropblock} to alleviate this problem. To a step further, we apply super-net guidance for better simulating the training process in multi-path schemes, where all weights are participated in training.

Compared with DARTS, each sub-graph in our method is learnt separately to ensure the same probability of being optimized, preventing Matthew effect from appearing. Furthermore, we can optimize $K$ sub-graphs simultaneously by sampling without replacement, which has stronger robustness than existing single-path sub-graph methods~\cite{cai2018proxylessnas,dong2019searching}, and thus avoids to fall into a local  optimum. In addition, our method can be considered as a building block of differentiable architecture search, and can be readily integrated into existing DARTS-based approaches, such as P-DARTS~\cite{chen2019progressive}, DARTS+~\cite{liang2019darts+}, Robust DARTS~\cite{Zela2019UnderstandingAR}, \textit{etc}. The main contributions of this work can be summarized as follows.

- We propose a Gumbel-TopK sampler that can sample $K$ sub-graphs without replacement in accordance with the available resource. By considering multiple mutually exclusive sub-graphs in the same iteration, our method delivers flexibility and fairness in the field of differentiable NAS.

- We tackle the proliferation of \textit{skip-connect} issue by applying Dropblock, which can serve as a training stabilizer that enables an effectively training of our multiple single-path sub-graphs searching. 

- We introduce a distillation strategy, called super-net guidance, for better optimizing our multiple mutually exclusive sub-graphs. The soft-labels generated by the super-net serve as an additional supervision for the sub-graphs.  It simulates training procedure in multi-path methods, and thus improves the training.

- We conduct extensive experiments to demonstrate the effectiveness of the proposed methods, and obtain a test error of 2.40\% on CIFAR10 and a top-1 error of 24.8\% on ImageNet, with a searching cost of 0.5 GPU days.

\section{Related Work}\label{related}

\noindent\textbf{Evolution Algorithm Based NAS.} EA-based methods~\cite{real2017large,elsken2018efficient,real2019regularized,liu2017hierarchical} leverage evolution algorithm to generate architecture. The focus in this field is how to encode the architecture and their mutation. Liu \emph{et al.} ~\cite{liu2017hierarchical} models the architecture using a hierarchical representation. Real \emph{et al.}~\cite{real2019regularized} studies a large CNN search space and proposes an aging evolution algorithm based on standard tournament selection. Their work is known as the first EA-based algorithm that beats handcraft architecture. \\

\noindent\textbf{Reinforcement Learning Based NAS.} This line of work~\cite{zoph2016neural,zoph2018learning,pham2018efficient} applies reinforcement learning to explore the architecture space. NAS~\cite{zoph2016neural} is the first work to use RNN meta-controller to generate the coding of architectures. Then the recurrent neural network can be optimized by the policy gradient or the proximal policy gradient under non-differentiable constraints, like FLOPS, the number of parameters, \textit{etc}. To ease the computational burden, NASNet~\cite{zoph2018learning} focuses on the building block of the network instead of the whole architecture. ENAS~\cite{pham2018efficient} speeds up the evaluation by sharing the weights among sub-graphs.\\

\noindent\textbf{Gradient Based NAS.} 
DARTS~\cite{liu2018darts} is the first work to introduce the gradient-based searching algorithm. They propose a continuous relaxation of the discrete search space. With this relaxation, the super-net weights and architecture parameters can be optimized in an alternative fashion, which reduces the searching time from months of GPU time to a single day. Nonetheless, it is still cursed by extravagant memory consumption. Besides, DARTS+~\cite{liang2019darts+} and DARTS-ES~\cite{zela2019understanding} show that DARTS tends to collapse and overfit due to the unbalanced competition between model parameters and architecture parameters. Other line of works like DARTS-~\cite{chu2020darts}, TE-NAS~\cite{zhang2021neural} and RLNAS~\cite{chen2021neural} use different performance indicators rather than validation loss to prevent searching collapse.
PC-DARTS~\cite{xu2019pc} reduces the memory usage by performing architecture search in a randomly sampled subset of operations. 

GDAS~\cite{dong2019searching} utilizes the Gumbel-Max trick to simulate the discrete sampling procedure during forward pass. 

Likewise, ProxylessNAS~\cite{cai2018proxylessnas} randomly selects one path from super-net in forward pass according to the given probability and adapts Binaryconnect~\cite{Courbariaux2015BinaryConnectTD} to back-propagate gradients. With less memory usage, ProxylessNAS can perform NAS directly on target dataset like ImageNet rather than on proxy dataset.

We are also aware of the works~\cite{chang2019differentiable,xie2018snas} where $K$ sub-graphs are optimized during training. Nonetheless, they obtain $K$ sub-graphs simply by running the Gumbel-SoftMax $K$ times, which is equivalent to sample the sub-graphs with replacement. DARTS-EGS~\cite{chang2019differentiable} aggregates all sub-graphs into a single graph to tackle the overlapping operations.  SNAS~\cite{xie2018snas} links their method with RL-based algorithms but ignores the potential of repeated update for same weight, which cause the unfair training of the super-net. In contrast to them, our differentiable sampler can output mutually exclusive sub-graphs, which covers the super-net to a larger extent in a single training iteration. \\

\noindent\textbf{Distillation.} Knowledge distillation ~\cite{hinton2015distilling} which transfers from a teacher network to a student one is turns out to be great effectiveness on NAS~\cite{liu2020search, li2020block,yu2020bignas,yu2020cyclic}. 

Most of these methods need to maintain an external network as a teacher. On the contrary, BigNAS~\cite{yu2020bignas} utilizes in-place distillation, which views the super-net itself as a teacher network, to eliminate extra training or post-processing steps for conventional NAS by greatly improving the accuracy of student network during searching phrase. Following BigNAS, we utilize the super-net as a teacher network to distill the sub-graphs networks.  

\section{Methodology}\label{method}
\begin{figure*}[tb]
    \centering
    \includegraphics[trim=0 190 50 100, clip,width=0.8\textwidth, height=15em]{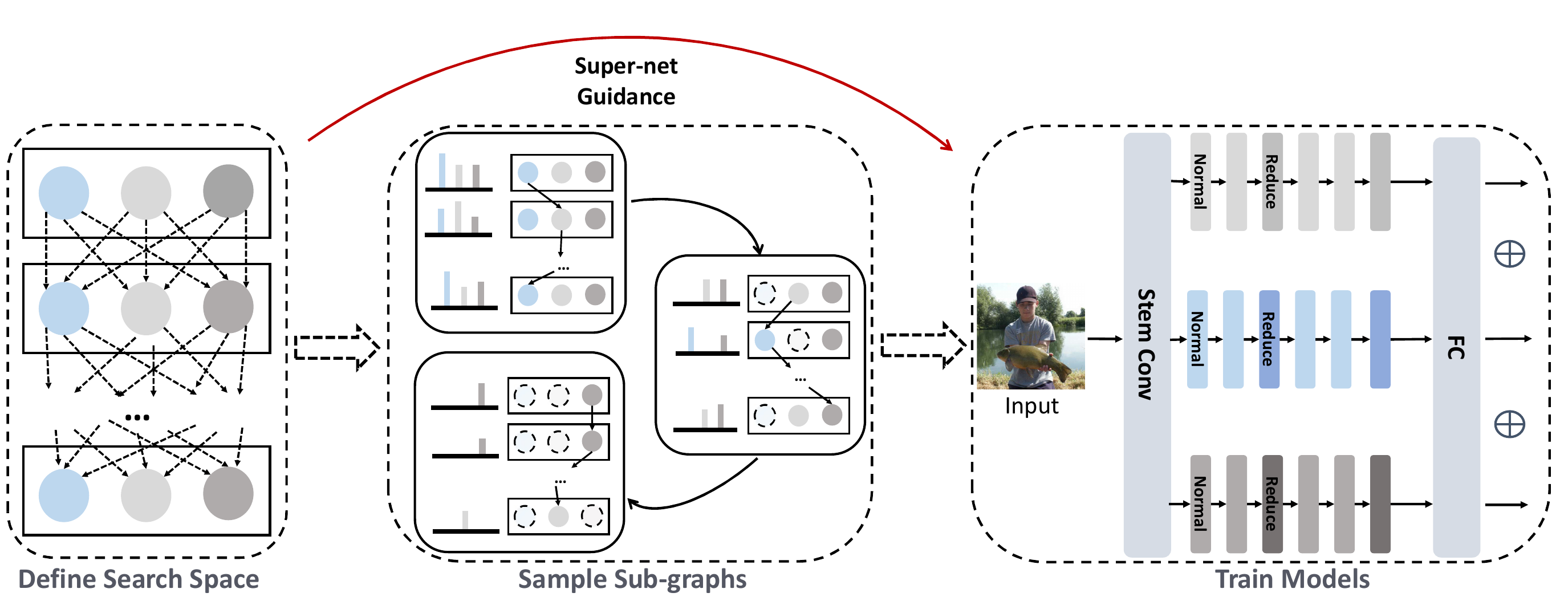}
    \captionsetup{font=small}
    \caption{\textbf{Illustration of the proposed algorithm.} For example, our Gumbel Top K samplers produces 3 exclusive cells from search space, which then are stacked to form 3 models. These models share the stem convolution layer and the final \textit{fc} layer. Each child model computes it own classification loss for the same training example and distillation loss with the super-net. The final loss is the summation of the loss of child models.}
    \label{fig:main}
\end{figure*}

In this section, we first briefly revisit DARTS~\cite{liu2018darts}, and then describe details of the proposed methods including three key components.\\

\noindent\textbf{Preliminary. } Our method is built on the framework of DARTS~\cite{liu2018darts}. We first revisit DARTS, which was designed for searching a robust cell structure, and then the full networks are constructed by stacking a series of the searched cells. A neural cell often consists of $N$ computational nodes represented by a directed acyclic graph (DAG), where each node $x_i$ is a latent representation (i.e., a stack of feature maps), and each edge $(i,j)$ from node $x_i$ to $x_j$ in the DAG indicates a bunch of feature transformation functions weighted by architecture parameters. Within a cell, each intermediate node is computed as a sum of all its predecessors.

The searching procedure in DARTS is formulated as a bi-level optimization problem. The model weights and architecture parameters are then updated in an alternative fashion by using a training set and a validation set, respectively. When the architecture parameters are learned, a full super-net can be pruned to derive a discrete architecture by maintaining the top-2 strongest predecessors for each intermediate node in the cell and keeping the top scoring operations.

In the setting of DARTS, all model weights in the super-net are learned simultaneously at each training iteration, which is a multi-path network that explores a large amount of memory cost. In addition, the training scheme of DARTS is different from the traditional approaches which use a single-path network. This may lead to an inaccurate estimation of the performance.\\

\noindent\textbf{Overview.} A single-path training scheme enjoys moderate memory usage as well as accurate approximation of the final model but may suffer from unstable optimization. It naturally comes up with a question that, is it possible to update multiple parts of the super-net while maintaining the merit of training the single-path sub-graph. To this end, we present a multiple single sub-graphs training scheme where multiple sub-graphs are sampled using an extension of a Gumbel sampler. We then propose to plug a DropBlock module replacing \textit{skip-connect} operations to stabilize the searching outcome. Moreover, we introduce a modified in-place distillation to remedy the gap between sub-graphs and super-net. Our methods can be summarized in Figure \ref{fig:main}.

\subsection{Mutually-aware Sub-graph Sampling}

In order to cover the whole super-net to a large extent, we develop a differentiable sampler that encourages the sampled models to have no overlap with each other. Besides, strict fairness can be imposed by uniformly sampling multiple sub-graphs without replacement within each iteration~\cite{chu2019fairnas}. Therefore, our method can meet such requirement by nature, and provides both flexibility and fairness, which set us apart from previous works. Our MSG-DAS can have a better trade-off between training efficiency and computational resource, and builds a bridge between the fairness and differentiable NAS. \\

\noindent\textbf{Gumbel-Softmax.}  Gumbel-Max Trick~\cite{yellott1977relationship} perturbs the architecture parameters $\bbb{\alpha}$ with Gumbel noise $\bbb{g}$, and then selects the operation having the highest score. GDAS~\cite{dong2019searching} and SNAS~\cite{xie2018snas} apply the Gumbel-Max Trick to each edge in a DAG, which allows them to sample a single-path sub-graph from a super-net. To embed the sampling procedure in the neural networks, it is usually relaxed to Gumbel-SoftMax~\cite{maddison2016concrete}, which can be formulated as,
\beq
\footnotesize
\bar{O}_{\text{GDAS}}\left(x_{i}\right) = \sum_{o \in \mathcal{O}}\frac{\exp \left((\alpha_{o} + g_o)/\tau\right)}{\sum_{o^{\prime} \in \mathcal{O}} \exp \left((\alpha_{o^{\prime}} + g_{o^{\prime}})/\tau\right)}\cdot o\left(x_{i}\right).
\eeq
where $g \sim \text{Gumbel} (0, 1)$, and $\tau$ is the temperature of SoftMax. The output of SoftMax approximates one-hot encoding when $\tau \rightarrow 0$. If $\tau \rightarrow \infty$, all elements in the logits vector are the same, and the approximated distribution is smooth. \\

\noindent\textbf{Extension to Gumbel-SoftMax.}
Our goal is to find a differentiable approach to sample multiple single-path sub-graphs. Moreover, these sub-graphs should be mutually exclusive, which requires that the same operation on each edge of super-net cannot be selected multiple times by a sampler. This casts the problem of producing multiple exclusive sub-graphs into the problem of sampling multiple items without replacement. The feasible domain for the next call of Gumbel-SofMax sampler should be updated by the previous selection. To be more specific, the previously selected elements ought to be removed from the available candidates:
\beq\label{eq:softTK}
\small
\begin{aligned}
&h_{o}^{k+1} = \frac{\exp \left((\alpha_{o} + g_o)/\tau\right)}{\sum_{o^{\prime} \in \mathcal{O}^{k+1}} \exp \left((\alpha_{o^{\prime}} + g_{o^{\prime}})/\tau\right)} \\
&\text{s.t.}~~\mathcal{O}^{k+1} = \mathcal{O}^{k} / \{o^1, o^2, \ldots, o^k\},~~\mathcal{O}^{1} = \mathcal{O}
\end{aligned}
\eeq

The main challenge is to carry out the pruning on the feasible space in a differentiable way. To this end, we propose to utilize a recent progress in top $K$ relaxation~\cite{plotz2018neural}. Let $\bbb{r} = (\bbb{\alpha} + \bbb{g})/\tau$ and $\hat{\bbb{h}}^{k}$ is the corresponding one-hot vector of $\bbb{h}^{k}$, we compute $\bbb{h}^{k+1}$ by setting the largest entry of $\bbb{h}^{k}$ to be negative infinity:
\beq\label{eq:r_update}
\small
\bbb{r}^{k+1} = \bbb{r}^{k}+\log \left(\bbb{1}-\hat{\bbb{h}}^{k}\right)
\eeq
The updated $\bbb{r}$ is then used in the computation for next selection. Details of the proposed sampler are summarized in Algorithm \ref{alg:soft_tk}.

\begin{algorithm}[!tb]
    \small
    \captionsetup{font=small}
	\caption{Gumbel Top-k Sampler}\label{alg:soft_tk}
	\begin{algorithmic}
	\STATE Input: architecture parameters $\bbb{\alpha}$, \# configuration $K$
	\STATE Output: $\hat{\bbb{h}}^{1}, \hat{\bbb{h}}^{2}, \ldots \hat{\bbb{h}}^{K}$
	\FOR{k=1:K}
        \STATE 1. Compute $\bbb{h}^{k}$ based on Eq.(\ref{eq:softTK})
        \STATE 2. Output one-hot encoding $\hat{\bbb{h}}^{k}$ from $\bbb{h}^{k}$
        \STATE 3. Update $\bbb{r}^{k}$ based on Eq.(\ref{eq:r_update})
    \ENDFOR
    \end{algorithmic}
\end{algorithm}

As illustrated in Figure \ref{fig:main}, we now can build the sub-graphs by stacking the exclusive cells sampled by our Gumbel TopK sampler. These sub-graphs share the same stem convolution layer and the final fully connected layer. The performance of sub-graphs at the same iteration only depends on their cells. The loss for optimization is computed by aggregating the losses of all models.

Furthermore, the proposed sampler allows for a flexible memory usage, which enables us to explore a larger batch size to further improve the performance. Besides, more training data within a mini batch can naturally provide a more stable update of the parameters~\cite{xu2019pc}. A good combination of the number of sub-graphs and batch size shows a promising direction for superior performance. For example, in our experiments, the test error can be further reduced from 2.81\% to 2.40\% on CIFAR10, by using a larger batch size (e.g., from B=64 to B=160).

\subsection{On Spread of \textit{Skip-connect} Issue}\label{skip-connect}

As mentioned in~\cite{liang2019darts+,chen2019progressive}, DARTS methods can easily lead to spread of \textit{skip-connect} operation, resulting in 
a significant shrinkage of the searched cells.
This inevitably limits the capacity of the derived networks, thus producing frustrated results. Therefore, the spread of \textit{skip-connect} has become a key issue for DARTS-based approaches.
In this work, we update $K$ single-path sub-graphs at each iteration. The \textit{skip-connect} can run separately in our multiple single-path sub-graphs, instead of working jointly with other operations, where the mixed operations can reduce the impact of \textit{skip-connect} operation. Besides, the \textit{skip-connect} operation often has a higher probability of being activated under our setting than the single-graph setting. These two factors make the spread of \textit{skip-connect} issue more significant in our algorithm. For example, a normal cell searched with $K=2$ can be filled with the \textit{skip-connect} when the training is stopped at 200 epoch. The parameter size of such architecture on CIFAR10 is merely 1.8 MB,  but having a large test error of 4.32\%.

This phenomenon is blamed to the collapse of model training caused by the rapid gradient passed through \textit{skip-connect}. A straightforward method to weaken the gradient is to randomly drop some elements in the gradient tensor, which can be achieved by applying Dropout~\cite{srivastava2014dropout}. However, Golnaz \emph{et al.} pointed out that Dropout works less effectively on convolutional layers than fully- connected layers. However, NAS is often performed on convolutional building blocks of the networks rather than the final FC layers, which inspired us to explore Dropblock~\cite{ghiasi2018dropblock} to alleviate the negative impact of \textit{skip-connect}.

Specifically, let $\tilde{m}$ be a Dropblock mask, and $\alpha_{id}$ be the corresponding architectural weight of the \textit{skip-connect}. Then the output can be written as, 
\beq
y = g(\alpha_{id})\cdot(x \cdot \tilde{m})
\eeq
where $g(\alpha_{id})$ denotes the Gumbel-SoftMax operation for $\alpha_{id}$.
The gradient with respect to $\alpha_{id}$ can be computed as,
\beq
\frac{\partial \mathcal{L}}{\partial \alpha_{id}}=\frac{\partial \mathcal{L}}{\partial y} \frac{\partial y}{\partial \alpha_{id}}=\frac{\partial \mathcal{L}}{\partial y} \frac{\partial g\left(\alpha_{id}\right)}{\partial \alpha_{id}}(x\cdot\tilde{m})
\eeq
Obviously, applying Dropblock on the output of \textit{skip-connect} is equivalent to dropping some gradients \textit{w.r.t} $\alpha_{id}$. The magnitude of gradients is then reduced effectively,  and the expectation of gradients now is,
\beq\label{eq:exp-gradient}
\begin{aligned}
\mathbb{E}\left[\nabla_{id }\right]&=\mathbb{E}\left[\frac{\partial \mathcal{L}}{\partial y} \frac{\partial g\left(\alpha_{id}\right)}{\partial \alpha_{id}}(x\cdot\tilde{m})\right] \\
&\approx \frac{\partial \mathcal{L}}{\partial \hat{y}} \frac{\partial g\left(\alpha_{id}\right)}{\partial \alpha_{id}}(\mathbb{E}[x]\cdot \mathbb{E}[\tilde{m}]) \\
&\approx \frac{\partial \mathcal{L}}{\partial \hat{y}} \frac{\partial g\left(\alpha_{id}\right)}{\partial \alpha_{id}}(\mathbb{E}[x]\cdot p) \\ 
s.t.~~&\tilde{m} \sim \text { Bernoulli }(p)
\end{aligned}
\eeq
where $\hat{y} = f(\alpha)\cdot x$, and we approximate the distribution of Dropblock with that of Dropout.

As implied by E.q. \ref{eq:exp-gradient}, the dropout probability should be sufficiently small, while ensuring a certain level of suppression on \textit{skip-connect}, to make it consistent with the derived model. We empirically found that training is stable in the early stage. Thus, we set the DropBlock ratio to be zero initially during training, and increase it in a step-wise manner overtime during training in our experiments.

\subsection{Gradient-stopped Super-net Guidance}
The whole super-net is updated in DARTS, while in MSG-DAS, only a fraction of weights are trained at each iteration. To bridge the gap between these two schedules, we propose a modified inplace distillation that transfers the information from super net to sub graph models in a GPU-memory-efficient manner. In the original inplace distillation~\cite{yu2019universally}, only the largest network is supervised by ground-truth labels. The other sub-networks are trained with soft-labels produced by the largest network. The memory consumption of this implementation is the same as DARTS since the training of super-net is required. 

\paragraph{Stopping Gradients of the Super-net.}
In our implementation, we only run the forward-pass of super-net to generate soft-label and no gradient is stored for saving the memory.

The training of super-net is driven by the update of sub-networks, and the training of sub-graphs are supervised by the ground truth labels. 

Specifically, the loss function of a sub-graph network consists of two ingredients, a conventional classification loss and a cross-entropy loss with soft-labels.
\beq\label{eq:loss}
\small
L^k =  L^k_{cls}+\lambda \sum_{c=1}^{C} p_c^{\text{super}} \log (p_c^{\text{sub-graph}})
\eeq
where $p_c^{\text{super}}$ denotes the soft-labels generated by super-net forwarding and $p_c^{\text{sub-graph}}$ represents the outputs of sub-graphs. 
The super-net can be viewed as a memory module where the historical updates of different paths are stored. Although we do not train the super-net explicitly, its output can still provide informative supervision.

The proposed distillation method is simple yet efficient. In contrast to the original version, it doesn't train the whole super-net at iteration and adds limited computational cost, while it still boosts the training like other knowledge distillation techniques.
Notably, since the super-net is not well trained at early stage, it does not provide much useful information. Hence, the guidance from super-net should play a small role at the beginning of searching. To cope with this, we adopt a linear schedule for the hyper-parameter of super-net guidance, starting from 0.

\section{Experiments}\label{expr}

\subsection{Datasets}

CIFAR10 include 50K training images and 10K test images, whose resolution is $32 \times 32$. The data from CIFAR10 is divided into 10 classes.

ImageNet~\cite{deng2009imagenet} is a popular large-scale benchmark for image classification, which contains 1K image categories, 1.2M  training images and 50K validation images. Following the mobile setting, we resize the images from ImageNet to $224 \times 224$ as input and the number of multiply-add operations in tested networks is restricted to be less than 600M.

\begin{table*}[!htb]
\begin{minipage}{0.52\linewidth}
\captionsetup{font=small}
\centering
\footnotesize
\setlength{\tabcolsep}{3pt}
\renewcommand{\arraystretch}{0.9} 
\caption{Classification result of our proposed algorithm and the baseline on CIFAR10. The searching costs are cited from their original papers.}
\begin{tabular}{ccccc}
\toprule
\multirow{2}[4]{*}{Architecture} & Test Err.(\%) & \multicolumn{1}{c}{Param} & \multicolumn{1}{c}{Search Cost} & \multicolumn{1}{c}{Search} \\
\cmidrule{2-2}      & CIFAR10 & \multicolumn{1}{c}{(M)} & \multicolumn{1}{c}{(GPU days)} & \multicolumn{1}{c}{Method} \\
\midrule
ResNet & 4.61  & 1.7   &   -    & manual \\
DenseNet-BC & 3.46  & 25.6  &   -    & manual \\
\midrule
NASNet-A & 2.65  & 3.3   & 1800  & RL \\
AmeobaNet-B & 2.55  & 2.8   & 3150  & evolution \\
ENAS  & 2.89  & 3.2   & 0.5   & RL \\
\midrule
ProxylessNAS & 2.08  &    -   & 4     & gradient \\
DARTS (1st order) & 3.00 $\pm$ 0.14 & 3.3   & 0.4   & gradient \\
DARTS (2nd order) & 2.76 $\pm$ 0.09 & 3.3   & 1     & gradient \\
GDAS  & 2.93  & 3.4   & 0.2   & gradient \\
SNAS (moderate) & 2.85 $\pm$ 0.02 & 2.8   & 1.5   & gradient \\
PC-DARTS & 2.57 $\pm$ 0.07 & 3.6   & 0.1   & gradient \\
P-DARTS & 2.50  & 3.4   & 0.3   & gradient \\
R-DARTS & 2.95 $\pm$ 0.21 &   -    & 1.6   & gradient \\
SDARTS-ADV & 2.61 $\pm$ 0.02 &    -   & 1.3   & gradient \\
Fair-DARTS & 2.54 $\pm$ 0.05 & \multicolumn{1}{l}{3.32 $\pm$ 0.46} &   -    & gradient \\
DARTS- & 2.59 $\pm$ 0.08 & 3.5 $\pm$ 0.13 & 0.4 & gradient \\
TE-NAS & 2.63 $\pm$ 0.0064 & 3.8 & 0.05 & pruning \\
\midrule
\textbf{MSG-DAS (B=128)} & 2.57 $\pm$ 0.05
  & 3.58 $\pm$ 0.09 & 0.5   & gradient \\
\textbf{MSG-DAS (B=160)} & 2.40  & 3.6   & 0.5   & gradient \\
\bottomrule
\end{tabular}%

\label{tb:cifar}
\end{minipage}
\quad
\begin{minipage}{0.5\textwidth}
\centering
\captionsetup{font=small}
\footnotesize
\setlength{\tabcolsep}{5pt}
\renewcommand{\arraystretch}{0.9} 
\caption{Top1 and Top5 errors of different architecture on ImageNet}
\begin{tabular}{ccccc}
\toprule
\multirow{2}[4]{*}{Architecture} & \multicolumn{2}{c}{Test Err.(\%)} & \multicolumn{1}{c}{Param} & MAC \\
\cmidrule{2-3}      & Top-1 & Top-5 & \multicolumn{1}{c}{(M)} & (M) \\
\midrule
MobileNet & 29.4  & 10.5  & 6.6   & 569 \\
MobileNet-v2 & 25.3  &   -    & 6.9   & 585 \\
ShuffleNet-v2 & 25.1  &   -    & 7.4   & 591 \\
\midrule
NASNet-A & 26.0  & 8.4   & 5.3   & 564 \\
MnasNet & 25.2  & 8.0   & 4.4   & 388 \\
AmeobaNet-C & 24.3  & 7.6   & 6.4   & 570 \\
MobileNet-v3-large & 24.8  &   -    & 5.4   & 219 \\
EfficientNet & 23.7  & 6.8   & 5.3   & 390 \\
\midrule
FBNET-A & 27.0  &   -    & 4.3   & 249 \\
SinglePath-NAS & 25.0  & 7.8   &   -    & - \\
ProxylessNAS (GPU) & 24.9  & 7.5   & 7.1   & 465 \\
DARTS & 26.7  & 8.7   & 4.7   & 574 \\
GDAS  & 26.0  & 8.5   & 5.3   & 530 \\
SNAS (mild) & 27.3  & 9.2   & 4.3   & 522 \\
DARTS- & 23.8  & 7.0 & 4.9 & 467 \\
TE-NAS & 26.2 & 8.3 & 6.3 & - \\
RLNAS & 24.4 & 7.5 & 5.3 & 581 \\
\midrule
\textbf{MSG-DAS} & 24.8  & 8.0   & 5.4   & 577 \\
\bottomrule
\end{tabular}%
\label{tb:imagenet}
\end{minipage}

\end{table*}

\begin{table*}[!hbt]
\begin{minipage}{0.5\linewidth}
\centering
\captionsetup{font=small}
\footnotesize
\setlength{\tabcolsep}{3pt}
\renewcommand{\arraystretch}{0.9}
\caption{Test errors on CIFAR10 with different components.}
\begin{tabular}{ccccc}
\toprule
Pseudo & Multi & Super-net & Test Err &  Param \\
\textcolor[rgb]{ .2,  .2,  .2}{Identity} & Graph & Guidance &  (\%) &  (M) \\
\midrule
      &       &       & 2.93  & 3.4 \\
\checkmark     &       &       & 2.91 $\pm$ 0.05 & 3.62 $\pm$ 0.2 \\
\checkmark     & \checkmark     &       & 2.76 $\pm$ 0.04 & 2.53 $\pm$ 0.03 \\
\checkmark     &       & \checkmark     & 2.77 $\pm$ 0.04 & 3.64 $\pm$ 0.18 \\
\checkmark     & \checkmark     & \checkmark     & 2.57 $\pm$ 0.05 & 3.58 $\pm$ 0.09 \\
\bottomrule
\end{tabular}%

\label{tb:ablation}
\end{minipage}
\quad
\begin{minipage}{0.5\linewidth}
\centering
\footnotesize
\setlength{\tabcolsep}{12pt}
\renewcommand{\arraystretch}{0.7}
\captionsetup{font=small}
\caption{Test errors with various super-net guidance weights}
\begin{tabular}{ccc}
\toprule
KL weight & Test Err(\%) &  Param \\
\textcolor[rgb]{ .2,  .2,  .2}{$\lambda$} & CIFAR10 &  (M) \\
\midrule
0     & 2.76 $\pm$ 0.04 & 2.53 $\pm$ 0.03 \\
1e-2  & 2.57 $\pm$ 0.05 & 3.58 $\pm$ 0.09 \\
1e-1  & 2.60 $\pm$ 0.0  & 3.84 $\pm$ 0.24\\
1     &  2.85 $\pm$ 0.11 & 3.94 $\pm$ 0.16\\
\bottomrule
\end{tabular}%

\label{tb:DML}
\end{minipage}
\end{table*}

\subsection{Implementation Details}

A complete experiment cycle consists of two steps, architecture search and architecture evaluation. We only perform architecture search for cell structure on CIFAR10. The final construction for the whole network varies with datasets. All tested networks are trained from scratch on target datasets. The experiments are conducted on a GTX 1080 Ti GPU without specification.\\

\noindent\textbf{Search Space.}~Our setup follows DARTS-like methods~\cite{liu2018darts,dong2019searching,chen2019progressive},
the operation space $\mathcal{O}$ in our experiments consists of the following operations: \textit{Zero}; \textit{Skip-connect}; \textit{3$\times$3 average pooling}; \textit{3$\times$3 max pooling}; \textit{3$\times$3 separable convolutions}; \textit{5$\times$5 separable convolutions}; \textit{3$\times$3 dilated convolutions}; \textit{5$\times$5 dilated convolutions}. The implementation of \textit{skip-connect} is modified as is discussed in Sec \ref{skip-connect}. 

A single cell has two input nodes, four computational nodes and a output node. The network is then formed by stacking 8 cells sequentially. The cell at level $l$ takes the outputs of previous two cells ($l-1$ and $l-2$) as inputs. Two reduction cells with stride 2 are located at 1/3 and 2/3 of the total depth of the network. The initial number of channels is 16. \\

\noindent\textbf{Searching Setup.}~We randomly partition the training data of CIFAR10 into training set $D_T$ and validation set $D_V$. The size of both sets are equal. We have conducted a series of experiment on the effect of $K$. For the sake of efficiency, we set $K=2$ for image classification tasks. The detail can be seen in Sec. \ref{subsec:discuss}. We search for a maximum of 240 epochs with batch size 128 without specification. Model weights $\bbb{w}$ are optimized using SGD with initial learning rate 0.025, a cosine annealing schedule, momentum 0.9 and weight decay $3 \times 10^{-4}$. All architecture parameters $\bbb{\alpha}$ are initialized as zero to ensure equal importance for all operations. We select Adam as the optimizer for $\bbb{\alpha}$, with initial learning rate 3$\times10^{-4}$, momentum (0.5, 0.999) and weight decay $10^{-4}$. The SoftMax temperature $\tau$ is initialized as 10 and linearly decays to 1.\\

\noindent\textbf{Evaluation Setup.}~We follow the practice of the previous works~\cite{liu2018darts} after obtaining the searched cell. We build a network of 20 cells and 36 initial channels for the evaluation on CIFAR10, which is slightly deeper than the one we used for searching on CIFAR10 to ensure a comparable model size as other baseline. The networks are trained on the whole training set for 600 epochs with batch size 64. Other hyper-parameters remain the same as the ones in search phrase. Some enhancements like cutout, path dropout with probability 0.2 and auxiliary towers with weight 0.4 are also considered. 

The network tested on ImageNet has 14 cells and the initial number of channels is 48. We train the network for 250 epochs with batch size 256, weight decay 3$\times10^{-4}$, initial learning rate 0.1 (decayed by a factor of 0.97 after each epoch) on eight GPUs. We also apply some training enhancements like label smoothing and auxiliary loss tower.

\subsection{Compared Methods}
Our implementation is based on the open source code of PC-DARTS\footnote{https://github.com/yuhuixu1993/PC-DARTS}. We set SNAS~\cite{xie2018snas}, GDAS~\cite{dong2019searching},Fair-DARTS~\cite{chu2019fair}, PC-DARTS~\cite{xu2019pc} and SDARTS-ADV~\cite{chen2020stabilizing} as our main opponents since they are in the same direction as us. GDAS considers a single-path sub-graph at each iteration. When $K=1$, our algorithm works similar to them. SNAS calls Gumbel sampler several times to produce sub-graphs, which may have overlapping parts. PC-DARTS activates a subset of operations in the super-net during search phrase. Our method can implement the similar idea by aggregating sub-graphs. Methods from different searching indicators~\cite{chu2020darts, zhang2021neural, chen2021neural} are also considered in our experiments. We also included the traditional handcraft models like ResNet~\cite{he2016deep}, DenseNet~\cite{huang2017densely}, MobileNet~\cite{howard2017mobilenets,sandler2018mobilenetv2,howard2019searching} and ShuffleNet~\cite{ma2018shufflenet}. EA based NAS~\cite{real2019regularized} and RL based NAS~\cite{zoph2016neural,tan2019mnasnet,pham2018efficient,tan2019efficientnet} are covered for comprehensive study.

\subsection{Results}

\noindent\textbf{Results on CIFAR10.}~We conducted the experiments three times with different random seeds (0,1,2). The results on CIFAR10 are revealed in Table \ref{tb:cifar}. MSG-DAS achieves a comparable error rate of 2.57\% on CIFAR10 in DARTS's search space, which exceeds our baseline DARTS and SNAS in both searching time and accuracy. The searching cost is slightly larger than GDAS, but the error is 0.53\% lower on CIFAR10. 

As we mentioned before, a larger batch size in architecture searching phrase is prone to produce a better result. Our method achieves 2.81\% test error on CIFAR10 with the same batch size as DARTS and GDAS (B=64). We obtain 2.61\% with same batch size as P-DARTS~\cite{chen2019progressive} (B=96). Our algorithm reaches 2.40\% with batch size 160 and outperform PC-DARTS (B=256).

Although the result of ProxylessNAS~\cite{cai2018proxylessnas} exceed  our MSG-DAS (top1 error is 2.08\% on CIFAR10), it costs  more search time (4.0 GPU days vs our 0.5 GPU days) and more parameters (5.7M VS our 3.6M). In addition, the search space of ProxylessNAS is different from ours, we believe that our MSG-DAS can also achieve considerable results when it transforms to the same space as ProxylessNAS. Analogously, our MSG-DAS can be easily integrated with it to further improve performance.\\

\noindent\textbf{Results on ImageNet.}~To test the generalization of the searched cell, we build networks for ImageNet with the architecture derived on CIFAR10.  We compare our results on ImageNet with the other methods in Table\ref{tb:imagenet}.  Architectures discovered on CIFAR10 by MSG-DAS outperform DARTS by a large margin in terms of classification performance, which demonstrate the transfer capability of the derived architecture. Additionally, MSG-DAS achieves lower test error than GDAS and SNAS (top1: 24.8\% VS 26.0\%, 27.3\%).  It should be noted that a well-designed search space tends to obtain superior performance~\cite{Yang2019NASEI}. Therefore, it is unfair to compare our method with a few recent methods which used the improved search spaces, such as EfficientNet~\cite{tan2019efficientnet} and MobilenetV3~\cite{howard2019searching}.\\

\subsection{Ablation Study}\label{subsec:discuss}

\noindent\textbf{Proposed Components.}~As shown in Table \ref{tb:ablation}, the modification of \textit{skip-connect} slightly increase the model size since less \textit{skip-connect} are selected in comparison to baseline model(GDAS). The test error can be further reduced to 2.77\% with the help of super-net guidance. Meanwhile, the introduction of multiple sub-graphs can also reduce the test error by 0.22\%. Our proposed method boosts the performance to 2.57\% with limited extra parameters compared to baseline. \\

\begin{figure}[!htb]
      \centering
      \captionsetup{font=small}
      \includegraphics[height=4cm]{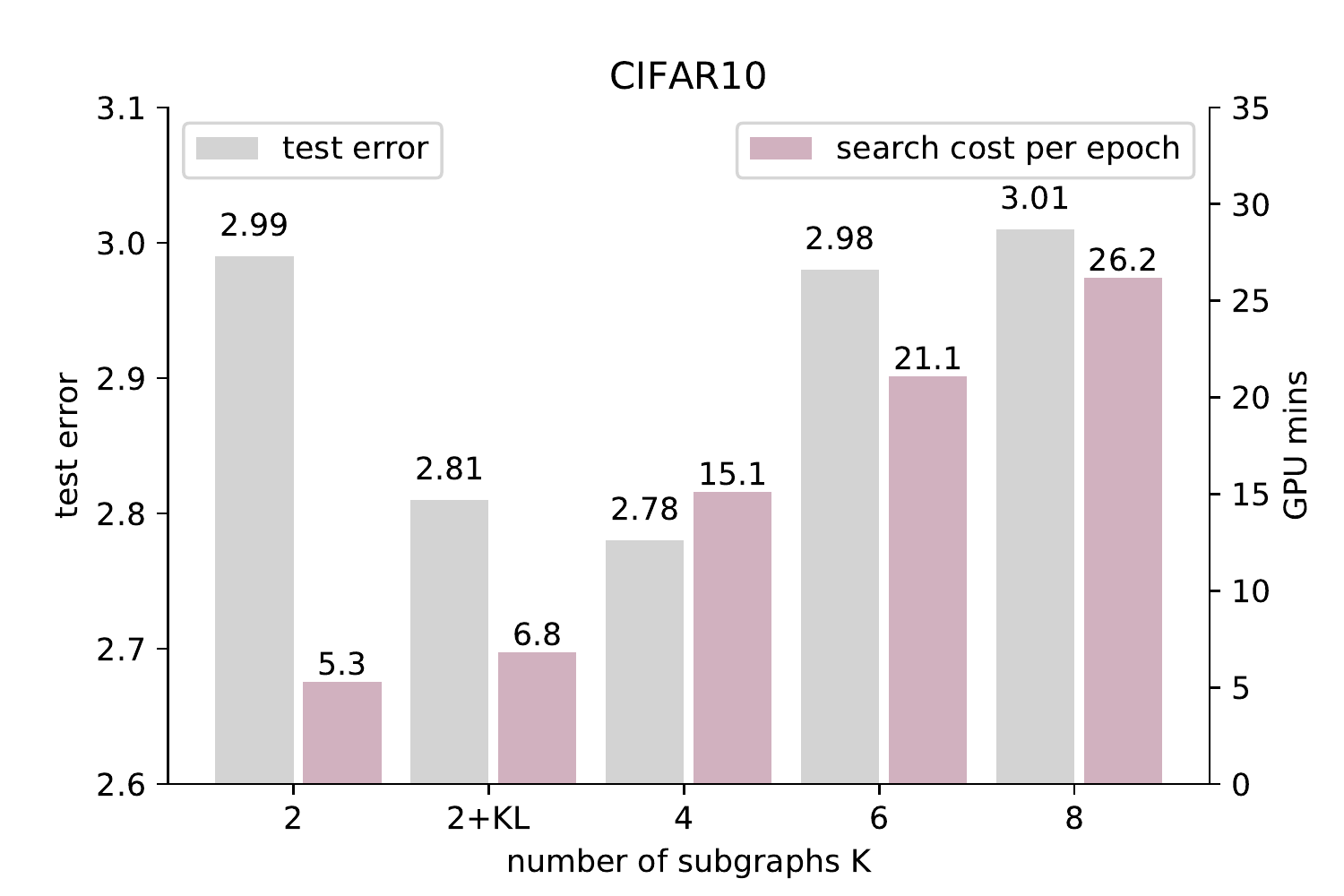}
      \caption{The test errors and searching costs on CIFAR10 under different number of sub-graphs. These models are searched with KL weight $\lambda=0$ without specification. The KL weight $\lambda$ for the model searched with mutual learning strategy is 0.01. The batch size for all setting is 64.}
      \label{fig:different_k}
\end{figure}

\noindent\textbf{Number of Sub-graphs.}~We plot the test error and search cost under different configuration in Figure \ref{fig:different_k}. More sub-graphs require more memory and the applicable batch size is smaller. Therefore, the experiments are conducted with the batch size 64 for fair comparison. Obviously, more computation time is needed with the increasing number of sub-graphs. With deep mutual learning strategy, only two sub-graphs ($K=2$) can get close to the best result (2.81\% vs. 2.78\%) with four-graphs($K=4$) but run twice faster. It indicates that our model with mutual learning strikes a good balance between accuracy and efficiency.

\noindent\textbf{Super-net Guidance.}~We then explore the influence of super-net guidance. The results are outlined in Table.\ref{tb:DML}. The performance peaks at 2.57\% when $\lambda = 0.01$. Besides, the test error deteriorates as the increasing $\lambda$. The optimization focuses on aligning the outputs instead of minimizing the error when the cross-entropy with soft-labels dominates the total loss and hampers the accuracy.

\section{Conclusion}\label{conclude}
We have presented a new searching framework (MSG-DAS) for differentiable architecture search where each searching sub-graph is single-path and mutually exclusive. We design a differentiable Gumbel-TopK sampler to produce  multiple  mutually exclusive single-path sub-graphs. Then we propose a Dropblock-Identity module to overcome the limitation of \emph{skip-connect} operation. Furthermore, we introduce a distillation strategy using cross-entropy to explore the additional supervision from super-net. The proposed MSG-DAS framework achieved comparable performance on both CIFAR10 and ImageNet, with a searching cost of 0.5 GPU days.

{\small
\bibliographystyle{ieee_fullname}

}

\end{document}